%% file: main.tex
\documentclass[conference]{IEEEtran}
\IEEEoverridecommandlockouts

\usepackage{cite}
\usepackage{amsmath,amssymb,amsfonts}
\usepackage{algorithmic}
\usepackage{graphicx}
\usepackage{textcomp}
\usepackage{xcolor}
\usepackage{balance}

\usepackage{algorithm}
\usepackage{pifont}
\usepackage{soul}
\usepackage{multirow}
\usepackage{makecell}

\def\BibTeX{{\rm B\kern-.05em{\sc i\kern-.025em b}\kern-.08em
    T\kern-.1667em\lower.7ex\hbox{E}\kern-.125emX}}
\begin{document}

\title{Towards 3D Acceleration for low-power Mixture-of-Experts and Multi-Head Attention Spiking Transformers
}

\author{
\IEEEauthorblockN{Boxun Xu\IEEEauthorrefmark{2}, Junyoung Hwang\IEEEauthorrefmark{3}, Pruek Vanna-iampikul\IEEEauthorrefmark{3}\IEEEauthorrefmark{4}, Yuxuan Yin\IEEEauthorrefmark{2}, Sung Kyu Lim\IEEEauthorrefmark{3}, Peng Li\IEEEauthorrefmark{2}\\
\IEEEauthorblockA{\IEEEauthorrefmark{2}Department of Electrical and Computer Engineering, University of California Santa Barbara, CA, USA}
\IEEEauthorblockA{\IEEEauthorrefmark{3}Department of Electrical and Computer Engineering, Georgia Institute of Technology, GA, USA \\}
\IEEEauthorblockA{\IEEEauthorrefmark{4}Department of Electrical Engineering, Burapha University, Chonburi, Thailand \\}}
}

\maketitle

\begin{abstract}
Spiking Neural Networks(SNNs) provide a brain-inspired and event-driven mechanism that is believed to be critical to unlock energy-efficient deep learning.  
The mixture-of-experts approach mirrors the parallel distributed processing of nervous systems, introducing conditional computation policies and expanding model capacity without scaling up the number of computational operation.
Additionally, spiking mixture-of-experts self-attention mechanisms enhance representation capacity, effectively capturing diverse patterns of entities and dependencies between visual or linguistic tokens.
However, there is currently a lack of hardware support for highly parallel distributed processing needed by spiking transformers, which embody a brain-inspired computation. This paper introduces the first 3D hardware architecture and design methodology for Mixture-of-Experts and Multi-Head Attention spiking transformers. By leveraging 3D integration with memory-on-logic and logic-on-logic stacking, we explore such brain-inspired accelerators with spatially stackable circuitry, demonstrating significant optimization of energy efficiency and latency compared to conventional 2D CMOS integration.

\end{abstract}


\section{Introduction}
Transformer models have significantly enhanced capabilities in language and vision tasks, gaining widespread adoption across diverse application domains \cite{TransformerImageRec2021,transformerText2Image21}. 
Compared to Convolutional Neural Networks (CNNs), the self-attention mechanism in transformers captures contextual relationships globally among all tokens in a long sequence, unifying global and local sequence details into a unified representation.
Within transformer architectures, Mixture-of-Experts (MoE), inspired by conditional computing mechanisms in neuroscience, has been widely adopted\cite{jiang2023mistral, clark2022unified}. MoE enables individual experts to specialize in learning specific features or storing task-specific knowledge, achieving significant performance improvements\cite{riquelme2021scaling, fan2022m3vit}. By leveraging a learnable gating function to efficiently route input tokens to appropriate experts, MoE decouples computational cost from model parameter size, achieving high scalability.

As the third generation of neural networks \cite{maass_snn_1997}, spiking neural networks (SNNs) exhibit a closer resemblance to biological neurons than conventional non-spiking artificial neural network (ANN) counterparts\cite{RoyNature:2019}. SNNs leverage powerful temporal coding, enable spatiotemporal computation through binary activations, and achieve ultra-low energy consumption on dedicated neuromorphic hardware platforms \cite{akopyan2015truenorth, davies2018loihi, PTB}.
Recent advancements in spiking neural network-based transformer models have successfully integrated self-attention mechanisms from traditional transformers into spiking neuron architectures. These models demonstrate superior performance over conventional network architectures \cite{spikformer, spikformer_tracking, zhu2023spikegpt, yao2023spike}, mirroring the trend observed in ANNs where vision transformers outperform ResNets. 

\begin{figure*}[ht]
    \centering
    \includegraphics[width=\textwidth, clip, trim={4.5cm 2.5cm 9cm 2.2cm}]{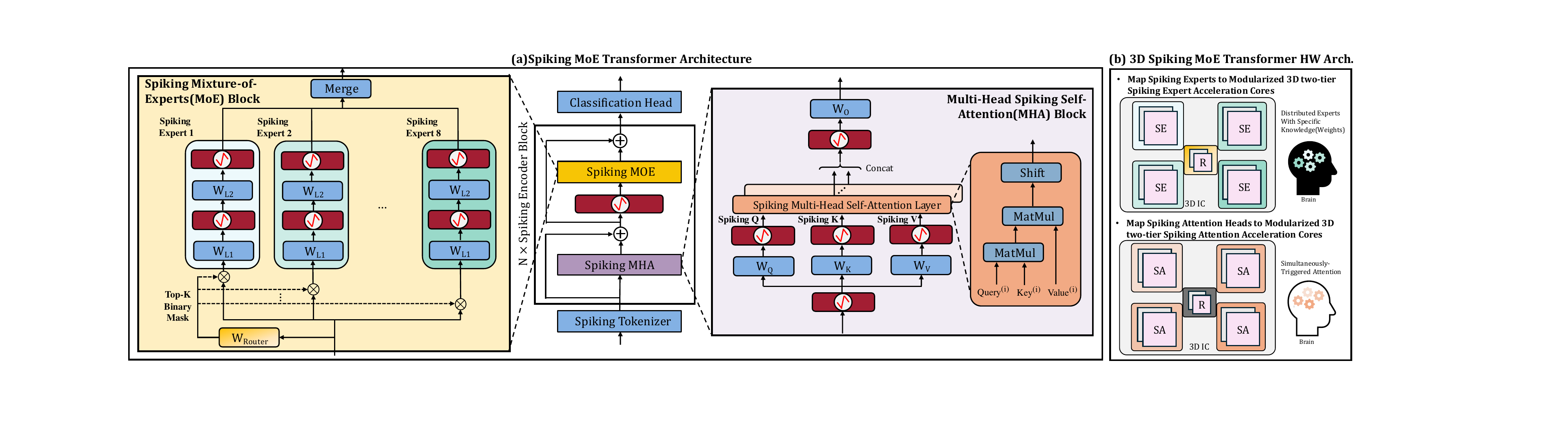}
    \caption{(a) Model Architecture of the Spiking Mixture-of-Experts (MoE) Transformer, comprising a core backbone with spiking MoE and spiking MHA (Multi-Head Attention) blocks. (b) Hardware architecture of the 3D Spiking MoE Transformer accelerator, featuring brain-inspired on-chip distributed experts and 3D simultaneously-triggered attention computation.}
    \label{fig:MoE_Arch}
\end{figure*}

Currently, there is a lack of dedicated hardware architectures designed for spiking transformers \cite{spikformer, spikformer_tracking, zhu2023spikegpt, yao2023spike}, particularly for spiking MoE transformers, presenting unique design challenges.
First, existing neuromorphic accelerators typically provide limited parallelism, constrained either to temporal dimensions \cite{Lee_ICCD_2020, PTB:HPCA:2022, yin2024loas} or spatial dimensions \cite{Spinalflow}. Moreover, these designs are predominantly tailored for accelerating spiking CNNs, making them ill-suited to address the computational demands and unique characteristics of large-scale spiking MoE transformer models.
Second, implementing brain-inspired algorithms on 2D silicon faces significant hardware overhead, including high memory access costs, complex routing requirements, and low computational density. These limitations prevent 2D designs from achieving the high computational efficiency of three-dimensional brain-like computing architectures because they defy the need for low power consumption in SNN.
Third, a naive 3D expert-by-expert implementation doesn't fully exploit the parallelism in spiking MoE transformers, leading to repeated weight loading and ignoring the distributed parallelism of spiking experts. 

\textbf{Challenges and Contributions}
In this work, we leverage face-to-face (F2F)-bonded 3D integration technology to design dedicated spiking Mixture-of-Experts (MoE) transformer accelerators incorporating spiking Multi-Head Attention (MHA) mechanisms. The proposed architecture employs both memory-on-logic and logic-on-logic configurations. 

\textbullet\space We present the first dedicated 3D accelerator architecture for spiking MoE transformers, efficiently exploring spatial and temporal parallelism weight reuse within modularized spiking experts and exploiting spiking expert parallelism for both MoE and MHA, supporting a scalable and efficient spike-based computation in MoE transformers.

\textbullet\space We explore the first 3D memory-on-logic and logic-on-logic interconnection schemes for the parallel distributed spiking MoE transformers to significantly reduce energy consumption and latency, thereby delivering highly efficient spiking neural computing systems with minimal area overhead. \\

Compared to the 2D CMOS integration, the 3D accelerators dedicated for spiking MoE transformers offer substantial improvements. For the spiking MoE and MHA workloads, it provides a 3\%-5.1\% increase in effective frequency, 39\%-41\% area reduction, 26.9\%-29\% memory access latency reduction and up to 14.4\% power reduction.

\section{Background}

\subsection{Spiking Neural Networks}\label{S2P1}

\subsubsection{Neuron Models in SNNs}
The spiking neuron models, as temporal activation functions, are widely used in SNNs. Leaky Integrate-and-Fire (LIF) models and Integrate-and-Fire (IF) models are commonly adopted. The LIF model simulates and mimics a neuron's response, exhibiting the following temporally discrete behaviors over multiple timesteps:
\begin{equation}\label{eqn_LIF_1}
V_{i}\left[t_{k}\right] = V_{i}\left[t_{k-1}\right] + \sum_{j\in RF}w_{ji}S_{j}[t_{k}] - V_{leak} \\
\end{equation}
\begin{equation}\label{eqn_LIF_2}
S_{i}\left[t_{k}\right] = \begin{cases} 
1 & \text{if  } V_{i}[t_{k}] > V_{th} \rightarrow V_{i}[t_{k}]=0\\
0 & \text{else          }                   \rightarrow V_{i}[t_{k}] = V_{i}[t_{k}]
\end{cases}
\end{equation}

\subsubsection{Spiking Multi-head Attention}
The Spiking Multi-Head Attention Mechanism splits the $D$ features of spiking $Q/K/V \in \mathbb{R}^{T \times N \times D}$ into $H$ slices, denoted as the number of attention heads, as $Q/K/V \in \mathbb{R}^{N \times T \times H \times d}$, where $D = H \times d$. Here, $T$ denotes the number of timesteps, and $N$ denotes the number of tokens. Each $Q_h \in \mathbb{R}^{N \times T \times d}$ and $K_h \in \mathbb{R}^{N \times T \times d}$ formulates $H$ spiking attention maps across $A \in \mathbb{R}^{H \times T \times N \times N}$. Each attention map is applied to the spiking value to generate the output as $Y = AV$.

\subsection{Mixture-of-Experts (MoE) Models}
Mixture-of-Experts (MoE) is a machine learning architecture that has gained traction for its high scalability. MoE models, leveraging a learnable routing network $W_r\in \mathbb{R}^{D_{in}\times E}$ to compute gating scores for $E$ experts, intelligently route input tokens to one or more of the most appropriate experts. These models are typically built on top of transformer-based models, where the traditional feed-forward network in each transformer layer is replaced with a combination of a gating network and multiple experts. To scale efficiently, MoE typically distributes experts across multiple GPUs \cite{fedus2022switch}, assigning one or more experts to each GPU while replicating non-expert parameters across all GPUs. However, the high energy consumption of GPUs conflicts with the low-power requirements of spiking neural networks (SNNs). To address this challenge, we propose a compact distributed MoE accelerator based on 3D integrated circuits, bridging the gap between scalability and energy efficiency \cite{zhao20243d}.

\begin{figure*}[ht]
    \centering
    \includegraphics[width=0.9\textwidth, clip, trim={2cm 1cm 2.5cm 1cm}]{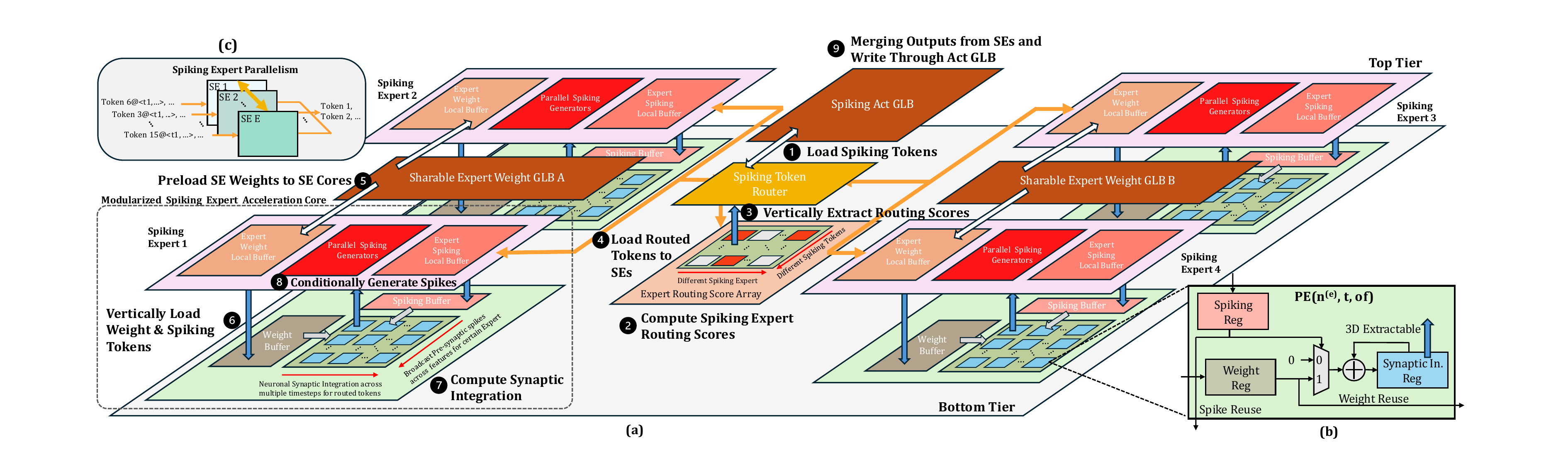}
    \caption{(a) Proposed 3D Architecture for processing Spiking Mixture-of-Experts layers: Two-tier 3D Partitioning and Dataflow (b) 3D Extractable PE Design (c) Expert Parallelism in Spiking MoE Transformers}
    \label{fig:3D_spiking_MoE}
\end{figure*}

\input{BackgrdNueroACC}

\section{Proposed 3D Spiking Mixture-of-Experts(MoE) Transformer Accelerators}\label{S3}

\subsection{Workload Processing of the Proposed Spiking MoE Layers}
Spiking Mixture-of-Expert(MoE) layers in spiking transformers encompass five core processing steps: 
\ding{192} Spiking (tokens) Conditional Routing(SCR), 
\ding{193} Spiking Synaptic Integration(SSI), 
\ding{194} Spiking Membrane potential Accumulation(SMA), 
\ding{195} Spiking Conditional Generation(SCG), and
\ding{196} Spiking Aligned Merging(SAM). 
Among these steps, \ding{192} and \ding{196} are inter-expert operations, which handle communication for routing and merging between different spiking expert pathways, while \ding{193}\ding{194}\ding{195} are intra-expert computations, responsible for the computationally intensive processing leveraging each individual spiking expert's knowledge.

Step \ding{192} processes the pre-synaptic activation $S_{in} \in \{0, 1\}^{N\times T\times D_{in}}$ to compute spiking expert scores $I\in\mathbb{R}^{N\times E}$, which quantify the importance of each spiking expert for a given spiking token. The top-K spiking experts are selected and assigned to the corresponding spiking tokens by routing. 
In step \ding{193}, each spiking expert processes the routed pre-synaptic spikes tokens for expert $e$, denoted as $S_{in}^{(e)}\in \{0, 1\}^{N_e\times T\times D_{in}}$. Each expert adapts pre-trained expert-specific weight $W^{(e)}\in\mathbb{R}^{D_{in}\times D_{out}}$, to compute the (post-)synaptic integration $X^{(e)}\in\mathbb{R}^{N_e\times T\times D_{in}\times D_{out}}$.
The step\ding{194} sequentially accumulates the synaptic integration of each neuron $i$ for each token $n$ at timestep $t$, denoted by $X_{n,i}^{(e)}[t]$ onto the membrane potential at timestep $(t-1)$, to update membrane potential $V_{n,i}^{(e)}[t]$.
Following this, each spiking expert, in step \ding{195}, performs conditional spike generation at each timestep as outlined in Equ.\ref{eqn_LIF_2}, generating $S_{out}^{(e)} \in \{0, 1\}^{N_e\times T\times D_{out}}$. 
Finally, as a merging function, \ding{196} aggregates the outputs from all experts $\{S_{out}^{(0)}, S_{out}^{(1)}, S_{out}^{(E-1)}\}$ to generate the aligned $S_{out} \in \{0, 1\}^{N\times T\times D_{out}}$ as outputs.



\subsection{3D Integrated Mixture-of-Expert (MoE) Spiking Transformer Accelerators}

In our proposed 3D integrated two-tier Spiking MoE accelerator design, as shown in Fig.~\ref{fig:3D_spiking_MoE}(a), we assemble multiple modularized Spiking Expert (SE) cores on a single chip and place two sharable expert-weight Global Buffers (GLBs) between the distributed SEs on the top tier. A centralized spiking activation GLB is also placed on the top tier, managing the spiking workload. Additionally, a two-tier spiking token router is placed between the four SE cores to enable the spiking expert parallelism as shown in Fig.~\ref{fig:3D_spiking_MoE}(c) by handling \ding{192} and \ding{196}.
Each SE core is responsible for performing computations corresponding to \ding{193}+\ding{194}+\ding{195} in parallel.

As illustrated in Alg.~\ref{alg:KF-S-MoE}, the router adapts $W_r$ to compute expert scores and based on the scores to route spiking tokens to distributed spiking expert cores. Each spiking expert $e$ receives routed workload $S_{in}^{(e)}$ and applies expert-specific weight $W^{(e)}$ locally to generate $S_{out}^{(e)}$ in parallel, which is then merged and written back to the Activation GLB.
Correspondingly, within the spiking token router, an expert routing score array is placed at bottom tier which computes the routing score for tiled tokens and the scores can be vertically extracted to the top tier. Then, the router selects the top-K spiking experts for each token, and routes the packed spiking tokens to multiple spiking experts. 
Within each SE core, the systolic PE array core placed at the bottom tier and a dedicated spiking generator core is placed at the top, executing kernel-fused operations of \ding{193}+\ding{194}+\ding{195}.

\input{MoE_algorithm}

\begin{figure*}[ht]
    \centering
    \includegraphics[width=0.9\textwidth, clip, trim={0.5cm 0.1cm 2.2cm 2cm}]{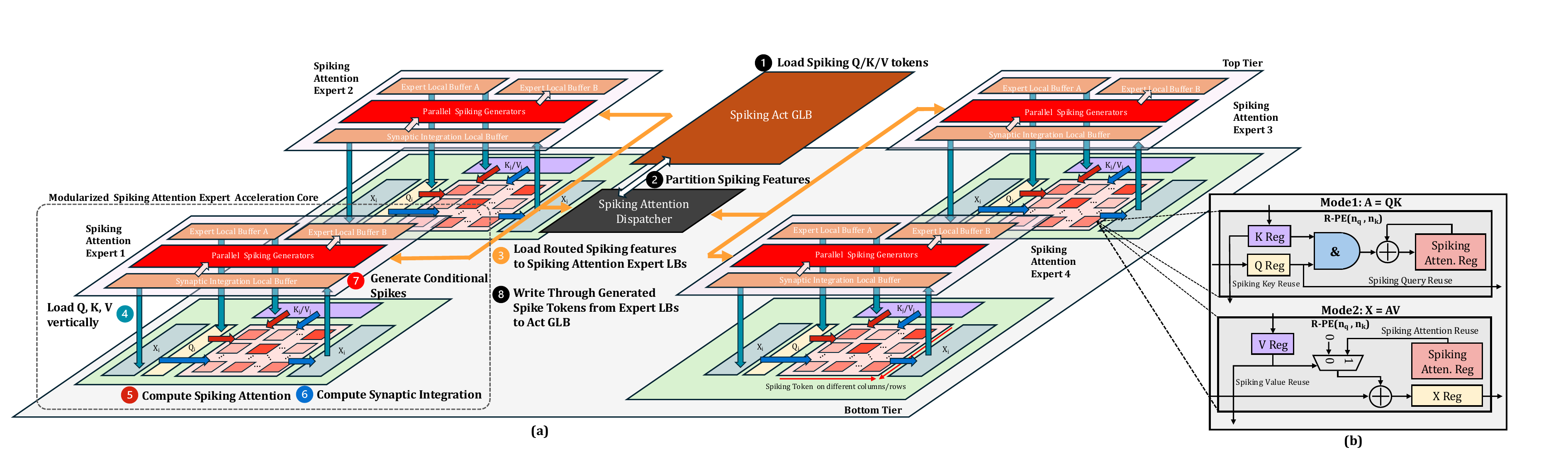}
    \caption{(a) Proposed 3D Architecture for processing Spiking Multi-head Attention layers (b) Reconfigurable PE Design}
    \label{fig:3D_spiking_MHA}
\end{figure*}

\input{MHA_algorithm}

As illustrated in Fig.~\ref{fig:3D_spiking_MoE}(a),  the optimized dataflow proceeds as follows: 
Pre-synaptic activation tiles ($S_{in}$) and expert routing weights ($W_r$) are first vertically loaded from the Spiking Act GLB and Weight GLB at the top tier to the bottom tier (Fig.~\ref{fig:3D_spiking_MoE} \ding{182}). 
Next, the expert routing array computes scores of $E$ experts for $N$ tokens in parallel, with the routing scores extracted to the top-tier router (\ding{183} and \ding{184}). 
The router then assigns spiking tokens to expert cores based on top-K selections, while the corresponding expert weights $W_e$ are preloaded into SE cores from the Weight GLBs (\ding{185} and \ding{186}). 
Subsequently, within each SE core, spiking tokens and weights are vertically loaded from local buffers to the bottom tier (\ding{187}), and synaptic integration for $D_{out}$ neurons over $N_e$ tokens and $T$ timesteps is performed within the dense spatiotemporal systolic array (\ding{188}). 
The integrated results are then extracted into spiking generators which compute membrane potentials and conditionally generate output spikes(\ding{189}), and the router   write the aligned results back to the Act GLB (\ding{190}).  
When processing assigned workload for an expert, the dense array within each modularized spiking expert core unrolls the workload with extreme fine spatiotemporal granularity. It maps 1-bit spiking activities to different columns, while multi-bit weights of output features are mapped to different rows.
Multi-bit weights are propagated horizontally, being reused across tokens and timesteps; 1-bit spiking activities $S_{in}^{(e)}$ propagate vertically and are reused across output neurons. 
As shown in Fig.~\ref{fig:3D_spiking_MoE}(c), each PE is designed with a synaptic integration-stationary approach. Synaptic integrations stored within PEs are 3D-extractable via dedicated readout ports, enabling spiking generators in the top tier to access data efficiently, thereby improving computation density.



\section{Proposed 3D Spiking Multi-Head Attention(MHA) Accelerators}\label{S4}

\subsection{Workload Processing of the MHA Layers}\label{S4P1}
The computation of spiking MHA layers is another bottleneck and has several key operations: \ding{192} the computation of spiking attention maps ($A = QK^T$), \ding{193} attention-weighted synaptic integration ($X=AV$), which provides inputs to a set of LIF neurons for generating the final binary spike-based attention output, \ding{194} membrane potential accumulation of these LIF neurons, and \ding{195} conditional generation of the LIF neuron output spikes as the final attention output. 
In operation \ding{192}, the spiking query $Q$ and spiking key $K$, initially shaped as $\mathbb{R}^{T\times N\times D_{in}}$, are divided into $\mathbb{R}^{T\times N\times H\times d}$. Here, $T$ represents the number of timesteps; N denotes the number of tokens; $H$ and $d$ indicate the number of self-attention heads and the number of features per head, respectively. A spiking attention map $S \in \mathbb{R}^{T\times H\times N\times N}$ is computed for each head at each timestep. For instance, the spiking attention map at $t$-th timestep for $h$-th self-attention head results from the binary matrix multiplication of the spiking query and key at the specific head and timestep. 
In \ding{193}, the attention-weighted synaptic integration is executed for each head at each timestep. The spiking attention map $A$, serving as the attention weights, is combined with the spiking value $V$, shaped in $\mathbb{R}^{T\times N\times H\times d} $ to compute attention-weighted synaptic integration, denoted by $X$ shaped as $\mathbb{R}^{T\times N\times H\times d}$.

\subsection{3D Integrated Spiking Multi-Head Attention Transformer Accelerators}
In the MHA mechanism described above, the computations of different heads can be neatly separated and processed by different spiking attention expert cores. The 3D MHA accelerators partition along with features and dispatch the partitioned features into different spiking attention experts to compute spiking outputs in parallel, and then neatly concat them and write through to the spiking Act GLB, as illustrated in Alg.~\ref{alg:KF-S-ATTN}. As illustrated in Fig.~\ref{fig:3D_spiking_MHA}(a), \ding{182} loads spiking Q/K/V tokens to a spiking attention dispatcher on the top tier. The spiking dispatcher neatly partitions spiking features, required for different attention heads, and routes the spiking features to distributed spiking attention expert local buffers(LBs) in \ding{183} and \ding{184}.

Each modularized spiking attention expert acceleration core performs \ding{192}-\ding{195} mentioned in Sec.~\ref{S4P1}. This acceleration performs kernel-fused spiking attention operations as mentioned in Alg.~\ref{alg:KF-S-ATTN} by adapting reconfigurable PEs in Fig.~\ref{fig:3D_spiking_MHA}(b).
In step \ding{185}, spiking $Q^{(h)}$, $K^{(h)}$ and $V^{(h)}$ are vertically loaded into buffers at the bottom tier. Then, $Q^{(h)}$ and $K^{(h)}$ are streamed into the array horizontally and vertically(\ding{186}), to compute spiking attention and store the attention matrix $A^{(h)}$ in internal registers within the reconfigurable array, which avoids a data movement of multi-bit attention. In \ding{187}, $V^{(h)}$ is streaming from top to down again to accumulate the attention-weighted synaptic integration that streams out at the right side of the PE array.  Subsequently, in step \ding{188}, the synaptic integration at the timestep is accumulated onto the membrane potential of the previous timestep stored on the top tier to generate conditional spikes in parallel spiking generators. Finally, the dispatcher writes through the generated spike activities back to spiking Act GLB in \ding{189}.

\section{Evaluations}\label{S6}
\subsection{Experiment Settings}
\textbf{Software Settings} 
We evaluated the spiking MoE transformer models on CIFAR10 and CIFAR100\cite{krizhevsky2014cifar}. The models utilize 8-bit quantized weights and 16-bit quantized synaptic integration. Each expert is allocated the same amount of weight parameters and trained for 100 epochs. We set feature size, patch size, head size and batch size to 128, 4$\times$4, 16 and 512, respectively. We set top-1 routing to keep an approximate number of operations with increasing the number of experts. We demonstrate that as the number of spiking experts scale up, the performance is improved significantly in Tab.~\ref{tab:MoE}.
\input{MoE_perf}

\textbf{Hardware Settings}
In this work, we use the commercial 28nm PDK to implement both 2D and 3D F2F designs. The 2D design consists of 6 metal layers, while the 3D design features a double metal stack of the 2D design, with the F2F bond pitch varying from 0.5$\mu m$ to 1$\mu m$. We use Synopsys Design Compiler to synthesize the RTL to a gate-level netlist and Cadence Innovus to perform physical synthesis. We utilize the pin-3D~\cite{PIN3D} flow, where the top and bottom dies are iteratively optimized across placement, CTS, routing, and sign-off stages.  
For memory, we utilize SRAM modules generated by a commercial memory compiler for various global buffers, local buffers, and other storage functions within our system architecture. On the top tier, for MoE accelerators, we adapt two 8K$\times$128b SRAM units employed for the Weight GLB A and Weight GLB B, which store shareable expert weights. For both MoE and MHA accelerators, an additional 8K$\times$128b SRAM unit is adapted to manage and store spiking activations. Within each modularized expert core, 3K$\times$128b SRAM units are employed for Activation, Weight, and Synaptic Integration Local Buffers. Additionally, smaller 96$\times$128b SRAM macros are allocated for the Query (Q) buffer, Key/Value (K/V) buffer, and Spiking (S) buffer on the bottom tier. Two 96$\times$256b SRAM macros are configured to serve as extended X buffers.  
The memory macro placement is determined based on the architecture information in Fig.~\ref{fig:3D_spiking_MHA} (for the MHA accelerator) and Fig.~\ref{fig:3D_spiking_MoE} (for the MoE accelerator). 
In MoE accelerators, we use a $16\times 128$ PE array with a $16\times 8$ routing score computing array; For the modularized spiking attention expert core, the size of reconfigurable attention array is $16\times 16$.

\subsection{Performance Comparison between 2D and 3D}
\subsubsection{Layout Comparision between 2D and 3D}
\begin{figure}[ht]
    \centering
    \includegraphics[width=0.45\textwidth, clip, trim={0cm 0cm 0cm 0cm}]{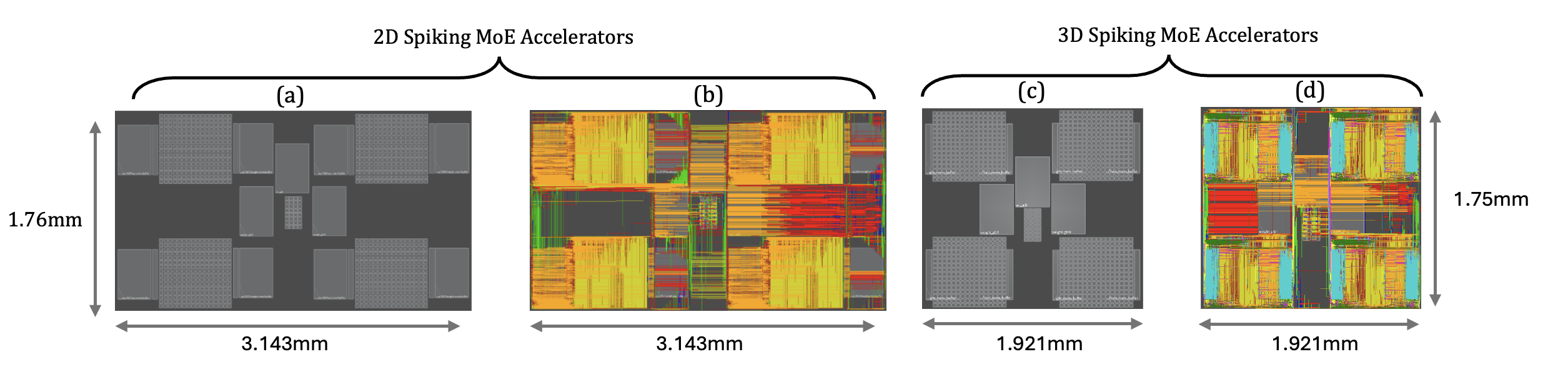}
    \caption{The placement and layout comparison of 2D and 3D spiking MoE accelerators.}
    \label{fig:MoE_layout}
\end{figure}

\begin{figure}[ht]
    \centering
    \includegraphics[width=0.45\textwidth, clip, trim={0cm 0.5cm 0.5cm 1cm}]{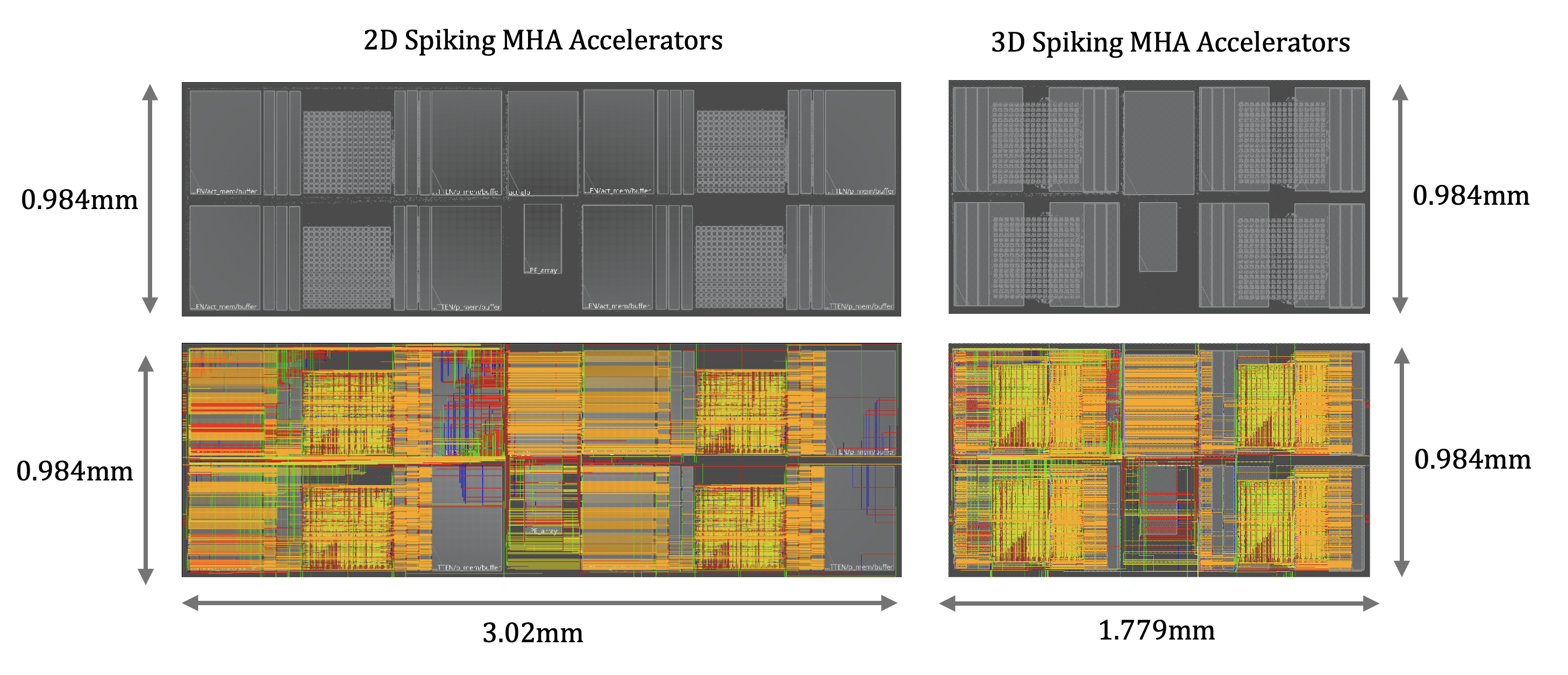}
    \caption{The placement and layout of 2D and 3D spiking MHA accelerators.}
    \label{fig:MHA_layout}
\end{figure}

In Fig.~\ref{fig:MoE_layout} and Fig.~\ref{fig:MHA_layout}, the placement and layout differences between the 2D and 3D designs of spiking MLP accelerators and spiking self-attention accelerators are presented. 
In Fig.~\ref{fig:MoE_layout}(a)(b), the 2D design occupies $1.76 \, \text{mm} \times 3.413 \, \text{mm}$, while the stacked 3D spiking design occupies $1.921 \, \text{mm} \times 1.75 \, \text{mm}$. On the top tier, the W GLBs and Act GLB are placed between expert cores, as shown in Fig.~\ref{fig:MoE_layout}(c)(d), while modularized Expert LBs and spiking generators are placed at the edges. Within each expert module, the weight and spike buffers are placed on the edge, and the spiking spatiotemporal array is positioned below the spiking generators. 
In Fig.~\ref{fig:MHA_layout}, the 2D design of the spiking attention accelerator occupies $0.984 \, \text{mm} \times 3.02 \, \text{mm}$, while the stacked 3D spiking design occupies $0.984 \, \text{mm} \times 1.779 \, \text{mm}$. In the 3D MHA design, the Act GLB is stored on the top tier, while other memories are stacked at the edges.

\subsubsection{PPA Comparison between 2D and 3D}

The performance comparison between 2D and 3D implementations of modularized spiking expert MHA and MoE accelerators, as shown in Tab.~\ref{tab:MoE_PPA_table}, demonstrates improvements across multiple metrics through 3D integration. In terms of performance, the 3D structure operates at higher effective frequencies of 2.24 GHz and 1.74 GHz for MHA and MoE, respectively, compared to their 2D counterparts at 2.13 GHz and 1.69 GHz, while achieving a reduced area footprint.

The area results of the 3D implementation show 39\% and 41\% reductions in MHA and MoE designs, respectively. Specifically, the MHA design area decreases from 5.53 $mm^2$ to 3.36 $mm^2$, while the MoE design reduces from 2.97 $mm^2$ to 1.75 $mm^2$. This area reduction is achieved while maintaining similar cell counts, indicating effective vertical integration without functionality loss. Power consumption results also show improvements in the 3D structure. In the MHA implementation, total power consumption decreases from 912 mW to 896 mW; the MoE design shows a larger improvement, with total power consumption reducing from 6989 mW to 5983 mW, representing a 14.4\% reduction. These improvements are observed across internal, leakage, and memory access power components.

\input{PPA}
\input{memory_access}
\input{wire}

\subsection{Analysis of Memory Access}
Memory access efficiency is a critical factor in determining the overall performance of neural network accelerators. Our analysis focuses on the memory access characteristics of both 2D and 3D implementations of MHA and MoE designs, as shown in Tab.~\ref{tab:MoE_PPA_table}. 
The 3D structure demonstrates improved memory access performance compared to its 2D counterpart. In the MHA design, memory access latency decreases from 160 ps to 112 ps, representing a 30\% reduction. Similarly, the MoE implementation shows a 15\% reduction in memory access latency, from 202 ps to 172 ps. These reductions in latency can be attributed to the shortened interconnect distances achieved through vertical integration. The improved memory access efficiency is also reflected in power consumption. The MHA design shows a 29\% reduction in memory access power, decreasing from 6.23 mW to 4.41 mW. Similarly, the MoE design achieves a 26.9\% reduction, from 7.11 mW to 5.2 mW. These reductions in both latency and power consumption demonstrate how 3D integration can optimize memory access patterns in neural network accelerators.

\subsection{Average Wirelength of 2D and 3D designs}
The wirelength comparison between 2D and 3D implementations, as shown in Tab.~\ref{tab:wire}, demonstrates notable reductions through 3D integration. For the single expert implementation, the MHA design shows a slight decrease from 0.621 m to 0.616 m, while the MoE design achieves a more significant reduction from 2.178 m to 1.959 m. The improvement becomes more pronounced in the 4-modularized expert systems, where the MHA design's wirelength reduces from 3.654 m to 3.290 m, and the MoE design shows a substantial decrease from 11.352 m to 9.816 m.
The reduction in wirelength directly contributes to the improved performance metrics observed in both designs. This is particularly evident in the hierarchical memory access characteristics detailed in Tab.~\ref{tab:Mem_Access_MoE_MHA}.

At the Global Buffer (GLB) level, the MHA design shows reduced activation latency from 220 ps to 209 ps, with power consumption decreasing from 10.9 mW to 7.56 mW. The MoE design demonstrates even more substantial improvements, with activation GLB latency reducing from 148 ps to 117 ps and notable reductions in weight GLB latencies from 241/147 ps to 94/71 ps. The Local Buffer (LB) and buffer-level metrics show similar improvements, with particularly significant reductions in the activation and weight buffer latencies and power consumption. These improvements can be attributed to the optimized wirelength and more efficient signal routing achieved through 3D integration. 
These results demonstrate how the reduced wirelength in 3D designs contributes to enhanced system performance through improved signal propagation and reduced power consumption in memory access operations.


\section{Conclusions}
We present the first dedicated 3D acceleration for MoE and MHA spiking transformers, leveraging spatial and temporal parallelism, modularized 3D spiking expert acclerators, and efficient interconnections. Our 3D acceleration achieves significant improvements over 2D CMOS integration, delivering scalable and energy-efficient spiking neural computation with minimal hardware overhead. This work enables a practical deployment of large-scale spiking MoE transformers.


\balance
\bibliographystyle{IEEEtran}
\bibliography{ref1, ref2}

\end{document}

%% file: BackgrdNueroACC.tex
\subsection{Neuromorphic Accelerators}
Various neuromorphic accelerators have been developed to support SNN inference at multiple levels, including devices \cite{nebula}, circuits \cite{liu202430} and architectures \cite{davies2018loihi, Spinalflow, PTB}. 
While these accelerators contribute to the advancement of generic SNNs, they lack optimization for the scalable spiking MoE transformers.
The exploration of 3D integrated circuits (3D ICs) for SNN hardware primarily focuses on monolithic 3D (M3D) \cite{M3D_SNN} and face-to-face (F2F) bonding techniques \cite{F2F_SNN}, which are often based on traditional liquid state machine (LSM) architectures. 
Although these methodologies improve power-performance-area (PPA) metrics, they are limited by constraints in neuron count, which hinder effective dataflow optimization.

%% file: MoE_algorithm.tex
\begin{algorithm}[t]
\caption{Kernel-Fused Parallel Processing for Spiking Distributed Mixture-of-Experts Layers.}\label{alg:KF-S-MoE}
\begin{algorithmic}
\STATE {\bfseries Input:}  
    the number of spiking expert $E$,
    spiking activation $S_{in} \in \{0,1\}^{N\times T\times D_{in}}$,
    expert routing weight $W_r \in \mathbb{R}^{T\times D_{in}\times E}$,
    distributed expert weight $W \in \mathbb{R}^{E\times D_{in}\times D_{out}}$.
\STATE {\bfseries Output:}      
    spiking output $S_{out} \in \{0,1\}^{N\times T\times D_{out}}$.
    \STATE \textbf{Load} $W_r$ and $S_{in}$ to Expert Routing Score Array
    \STATE \textbf{Compute} expert scores $I$
    \STATE \textbf{Route} $S_{in}$ into $\{S_{in}^{(0)},... , S_{in}^{(E-1)}\}$ by selecting Top-K experts
\FOR{$e=0 \ \ \text{to} \ E-1$}
    \STATE \textbf{Preload} $W^{(e)}$ from W GLB0/1 into the LB. 
    \STATE \textbf{Compute} $S_{out}^{(e)}$ adapting modularized spiking expert accelerators
    \STATE \textbf{WriteThrough} $S_{out}^{(e)}$ to Spiking Act LB
\ENDFOR
\STATE \textbf{Merge} $\{S_{out}^{(0)}, ..., S_{out}^{(E-1)}\}$ to $S_{out}$ and \textbf{WriteThrough} to Act GLB
\end{algorithmic}
\end{algorithm}

%% file: MHA_algorithm.tex
\begin{algorithm}[t]
\caption{Kernel-Fused Parallel Processing for Spiking Multi-head Attention Layers}\label{alg:KF-S-ATTN}
\begin{algorithmic}
    \STATE {\bfseries Input:}  
        Spiking query $Q \in \{0,1\}^{T\times N\times H\times d}$,
        Spiking key $K \in \{0,1\}^{T\times N\times H\times d}$,
        Spiking value $V \in \{0,1\}^{T\times N\times H\times d}$,
        Kernel-fused operations of Spiking Attention core $SpikingAttention$.
    \STATE {\bfseries Output:}      
        spiking output $S_{out} \in \{0,1\}^{T\times N\times D}$.
    \STATE{\textbf{Partition} $Q/K/V$ loaded from spiking Act GLB into $\{Q^{(0)},...,Q^{(H-1)}\}, \{K^{(0)},...,K^{(H-1)}\}, \{V^{(0)},..., V^{(H-1)}\}$}
    \FOR{$h=0 \ \text{to}\ H-1$}
        \STATE{\textbf{Dispatch} $Q^{(h)}$,$K^{(h)}$, $V^{(h)}$ to Spiking Attention Core $h$}
        \STATE{$S_{out}^{(h)} \leftarrow SpikingAttention(Q^{(h)},K^{(h)}, V^{(h)})$}
    \ENDFOR
    \STATE{\textbf{Merge} $S_{out} \leftarrow Concat(S_{out}^{(0)},..., S_{out}^{(H-1)}) $}
    \STATE{\textbf{Write Through} $S_{out}$ to Spiking Act GLB}
\end{algorithmic}
\end{algorithm}

%% file: MoE_perf.tex
\begin{table}[H]
\centering
\caption{Performance comparison on CIFAR10 and CIFAR100.}
\begin{tabular}{c c c c}
\hline
\textbf{Dataset} & \textbf{\# Spiking Experts} & \textbf{Accuracy (\%)} & \textbf{\#Params} \\ \hline
\multirow{3}{*}{CIFAR10}  & 1 & 88.31  & 1.05M \\ 
                          & 4 & 91.63 & 2.67M \\ 
                          & 6 & 92.06 & 3.74M \\ \hline
\multirow{3}{*}{CIFAR100} & 1 & 62.00 & 1.07M \\ 
                          & 4 & 69.62 & 2.68M \\ 
                          & 6 & 69.97 & 3.75M \\ \hline
\end{tabular}
\label{tab:MoE}
\end{table}

%% file: PPA.tex
\begin{table}[H]
\centering
\caption{Performance comparison between 2D and 3D designs of 4-modularized spiking expert MHA and MoE accelerators}
\setlength{\tabcolsep}{4.5pt}
\begin{tabular}{ccccc}
\hline
\multirow{2}{*}{\# Modularized Experts = 4}   &   \multicolumn{2}{c}{MHA}  & \multicolumn{2}{c}{MoE} \\  
\cline{2-5}
                    & 2D   & 3D         & 2D   & 3D      \\ 
\hline
Effective Frequency ($GHz$) & 2.13 &  \textbf{2.24} &  1.69  &   \textbf{1.74}  \\ 
Area Footprint ($mm^2$)  & 5.53 & \textbf{3.36} & 2.97 & \textbf{1.75} \\ 
Number of Cells & 169046 & 167983 & 339846 & 339693 \\ 
\hline
Internal Power (mW) & 863 & 859 & 6777 &  5716 \\ 
Switching Power (mW) & 30 &  18 & 67 & 116  \\ 
Leakage Power (mW) & 19 & 18 & 144  &  111  \\
Total Power (mW) & 912 & \textbf{896} & 6989 & \textbf{5983} \\
\hline
Memory Access Latency (ps) & 160 & \textbf{112} & 202 & \textbf{172}\\
Memory Access Power (mW) & 6.23 & \textbf{4.41} & 7.11 & \textbf{5.2}\\
\hline
\end{tabular}    
\label{tab:MoE_PPA_table}
\end{table}

%% file: memory_access.tex
\begin{table}[]
\centering
\caption{The hierachical memory access overhead comparisons between 2D and 3D design of spiking MHA and MoE accelerator}
\setlength{\tabcolsep}{4.5pt}
\begin{tabular}{ccccc}
\hline
\multirow{2}{*}{\#Modularized Experts = 4}   &   \multicolumn{2}{c}{MHA}  & \multicolumn{2}{c}{MoE} \\  
\cline{2-5}
                                        & 2D   & 3D         & 2D   & 3D     \\ 
\hline
Activation GLB Latency ($ps$) & 220 &  209 &  148  &   117  \\ 
Activation GLB Power ($mW$)  & 10.9 & 7.56 & 2.36 & 1.84  \\ 
Weight GLB0/GLB1 Latency ($ps$) & - & - & 241/147 & 94/71 \\ 
Weight GLB0/GLB1 Power ($mW$) & - & - & 3.87/4.05 & 2.03/2.01 \\ 
\hline
Activation LB Latency ($ps$) & 24 &  16 &  68  &   19  \\ 
Activation LB Power ($mW$)  & 1.13 & 0.76 & 1.1 & 0.77  \\ 
Weight LB Latency ($ps$) & 82 & 26 & 77 & 18 \\ 
Weight LB Power ($mW$) & 0.46 & 0.1 & 0.47 & 0.09  \\ 
\hline
Activation Buffer Latency ($ps$) & 40 & 16 & 40 & 19  \\
Activation Buffer Power ($mW$) & 1.92 & 0.52 & 1.66 & 0.27  \\
Weight Buffer Latency ($ps$) & 28 & 26 & 77 & 18  \\
Weight Buffer Power ($mW$) & 1.01 & 0.17 & 1.50 & 0.39  \\
\hline
\end{tabular}    
\label{tab:Mem_Access_MoE_MHA}
\end{table}

%% file: wire.tex
\begin{table}[ht]
\centering
\caption{2D and 3D wirelength Comparison of MHA and MoE}
\begin{tabular}{ccccc}
\hline
\multirow{2}{*}{Design/Wirelength(m)} & \multicolumn{2}{c}{MHA} & \multicolumn{2}{c}{MoE} \\ 
\cline{2-5}
                          & 2D & 3D & 2D & 3D \\ 
\hline
1-Modularized-Expert-only & 0.621 & \textbf{0.616} & 2.178 & \textbf{1.959} \\
4-Modularized-Expert System    & 3.654 & \textbf{3.290} & 11.352 & \textbf{9.816} \\ 
\hline
\end{tabular}
\label{tab:wire}
\end{table}